\documentclass{article}

\usepackage{arxiv}

\usepackage[utf8]{inputenc}
\usepackage[T1]{fontenc}
\usepackage{hyperref}
\usepackage{url}
\usepackage{booktabs}
\usepackage{amsfonts}
\usepackage{nicefrac}
\usepackage{microtype}
\usepackage{amsmath}
\usepackage{amssymb}
\usepackage{graphicx}
\usepackage{float}
\usepackage{multirow}
\usepackage{array}
\usepackage{caption}
\usepackage{subcaption}
\usepackage{xcolor}

\graphicspath{{./}{./images/}}

\title{TRUST Agents: A Collaborative Multi-Agent Framework for Fake News Detection, Explainable Verification, and Logic-Aware Claim Reasoning}

\author{
  Aishwarya Gaddam \\
  George Mason University \\
  \texttt{agaddam3@gmu.edu}
  \And
  Gautama Shastry Bulusu Venkata \\
  George Mason University \\
  \texttt{sbulusuv@gmu.edu}
  \And
  Santhosh Kakarla \\
  George Mason University \\
  \texttt{skakarl3@gmu.edu}
  \And
  Maheedhar Omtri Mohan \\
  George Mason University \\
  \texttt{momtrimo@gmu.edu}
}

\begin{document}
\maketitle

\begin{abstract}
The large-scale spread of misinformation has increased the need for automated fact-checking systems that are not only accurate, but also interpretable, evidence-grounded, and robust enough to handle complex claims. In many realistic scenarios, identifying whether a statement is true or false is only one part of the problem. A trustworthy system must first identify which parts of a news article or user-provided statement are verifiable, then retrieve relevant evidence, compare that evidence with the extracted claims, reason over uncertainty, and finally generate an explanation that a human can inspect. Although recent large language models have shown promising performance across many reasoning tasks, direct single-model fact-checking pipelines still suffer from hallucinated justifications, weak evidence grounding, poor calibration, and difficulty handling compound claims that contain multiple logically connected propositions.

This paper presents \textbf{TRUST Agents}, a collaborative multi-agent framework for fake news detection, explainable fact verification, and structured reasoning over complex claims. The system is organized as a modular pipeline in which specialized agents perform claim extraction, evidence retrieval, verification, and explanation generation. In the baseline pipeline, a claim extractor combines named entity recognition, dependency parsing, and LLM-based extraction to identify factual claims. A retrieval agent then performs hybrid sparse and dense evidence retrieval using BM25 and FAISS. A verifier agent compares the claims against the retrieved evidence and produces verdicts with calibrated confidence. Finally, an explainer agent transforms the intermediate reasoning signals into a human-readable fact-check report with explicit evidence citations.

To address the limitations of flat claim verification and single-verifier reasoning, we further introduce a research-oriented extension of the framework. This enhanced pipeline adds a decomposer agent inspired by LoCal-style logical claim decomposition, a Delphi-inspired multi-agent jury containing multiple specialized verifier personas, and a logic aggregator that combines atomic verdicts according to conjunction, disjunction, negation, and implication structures. The goal of the research pipeline is to improve reasoning transparency and to better handle claims that are compositionally complex or causally structured.

We evaluate the baseline and research pipelines on the LIAR benchmark and compare them against fine-tuned BERT, fine-tuned RoBERTa, and a zero-shot LLM baseline. The results show that while supervised encoder baselines still outperform the multi-agent system in raw benchmark metrics, TRUST Agents provides substantial gains in modularity, interpretability, evidence transparency, and structured reasoning. The research pipeline is especially useful for decomposing multi-part claims and exposing verifier disagreement, though its performance is constrained by high abstention rates and the quality of retrieved evidence. Our findings suggest that retrieval quality and uncertainty calibration remain the central bottlenecks in evidence-grounded fact-checking, and that multi-agent architectures are most valuable when they are embedded inside a carefully designed retrieval-and-reasoning pipeline rather than used as standalone debate systems.
\end{abstract}

\section{Introduction}
The rapid spread of misinformation across digital platforms has made automated fact-checking an increasingly important research problem. News content, political statements, viral posts, and manipulated narratives can be produced and distributed at a scale that far exceeds the capacity of professional human fact-checkers. As this asymmetry grows, misinformation can influence elections, distort public discourse, fuel panic during crises, and undermine trust in institutions before corrections become visible. This has created a strong need for AI systems that can assist with identifying and verifying claims at scale.

However, fact-checking is fundamentally more complex than standard text classification. A sentence or article cannot simply be labeled as true or false in a vacuum. In realistic settings, many inputs contain multiple claims, only some of which are actually verifiable. Some claims are factual, while others are interpretive, rhetorical, or speculative. Even when a claim is clearly factual, determining its truth often depends on external evidence that may be fragmented, incomplete, conflicting, or outdated. As a result, a useful fact-checking system must do more than produce a label. It must isolate verifiable content, search for relevant evidence, compare claims against retrieved passages, evaluate confidence, and communicate its reasoning transparently.

Traditional supervised fake news detection systems often treat the task as end-to-end classification. Transformer-based models such as BERT and RoBERTa can achieve strong benchmark results when trained directly on labeled datasets, but they typically do not expose the reasoning process behind their predictions. In many cases, such models learn statistical patterns tied to the benchmark distribution rather than robust evidence-grounded verification strategies. This becomes especially problematic in high-stakes settings where explanations matter as much as predictions.

Recent large language models have expanded the design space for fact-checking systems. They can interpret language more flexibly, reason over claims, summarize retrieved passages, and generate detailed natural-language explanations. Yet naïvely applying a single LLM to verification raises several challenges. First, a single model may hallucinate evidence or generate justifications that sound plausible but are not traceable to actual sources. Second, many real-world claims are not atomic. They contain conjunctions, disjunctions, causal assertions, temporal relations, or embedded assumptions. A flat verifier often mishandles these claims because it tries to judge the entire statement as one unit. Third, different reasoning styles lead to different behaviors: a cautious verifier may abstain too often, while an aggressive verifier may overcommit under weak evidence. Finally, explanation quality depends heavily on whether the system reasons through evidence or merely generates a post hoc narrative.

This paper introduces \textbf{TRUST Agents}, a modular multi-agent framework designed to address these issues. The central idea is that automated fact-checking should be treated as a structured pipeline rather than a single monolithic decision. The process begins with claim extraction, followed by evidence retrieval, claim--evidence verification, and explanation generation. Each stage is assigned to a dedicated agent with a well-defined role, allowing the overall system to be more interpretable, easier to debug, and easier to improve incrementally.

The framework contains two variants. The first is a baseline pipeline that combines claim extraction, hybrid retrieval, verification, and explanation generation into an end-to-end verification system. The second is a research-enhanced pipeline that extends the baseline with a decomposer agent, a multi-persona verifier jury, and a logic-aware aggregation module. The goal of the enhanced system is not only to improve predictive performance, but also to handle compositional claims more faithfully and to surface internal reasoning structure that would otherwise remain hidden.

The contributions of this work are as follows. We present a modular multi-agent fact-checking architecture that integrates claim extraction, retrieval, verification, and explanation into a single workflow. We introduce a research extension that decomposes complex claims into atomic propositions and aggregates multi-agent verifier outputs using explicit logical structure. We evaluate both versions of the system on the LIAR benchmark and compare them against supervised encoder baselines and a zero-shot LLM verifier. Finally, we provide a detailed analysis of the system's behavior, especially the role of uncertainty, retrieval quality, evidence coverage, and reasoning transparency in multi-agent fact-checking.

\section{Related Work}
Research on misinformation detection and fact-checking spans several overlapping areas, including supervised truthfulness classification, evidence retrieval, natural language inference, explainable AI, and LLM-based reasoning systems. Earlier work in fake news detection focused primarily on predicting whether an input statement or article was true or false based on text, speaker metadata, propagation patterns, or source-level information. These systems were useful as benchmark classifiers, but they often lacked explicit evidence grounding and offered limited interpretability.

The release of datasets such as LIAR enabled more systematic evaluation of truthfulness prediction models. Transformer-based encoders such as BERT and RoBERTa quickly became strong baselines because of their ability to learn contextual representations from short political claims. However, these models are still primarily label predictors rather than evidence-based reasoners. They often perform well by exploiting latent statistical regularities in benchmark datasets, but they do not inherently retrieve evidence or explain why a claim should be judged as true or false.

Another major direction in the literature involves evidence-based fact verification. In these systems, a claim is first linked to an external evidence corpus such as Wikipedia, news sources, or curated knowledge repositories. Sparse retrieval methods such as BM25 are widely used because they provide strong lexical matching, while dense retrieval methods provide semantic flexibility through embedding-based search. Hybrid retrieval systems combine both approaches and have become increasingly popular in verification pipelines because they improve evidence recall under lexical variation.

More recent work has explored the use of large language models for claim verification, explanation generation, and tool-augmented reasoning. LLMs can perform decomposition, summarize evidence, and emulate natural-language inference behavior over retrieved passages. However, they also suffer from hallucinations and weak calibration, especially when evidence is incomplete. This has motivated research into tool-using agents, retrieval-grounded pipelines, and collaborative multi-agent systems.

Multi-agent reasoning has become especially relevant for difficult reasoning tasks. Instead of relying on a single verifier prompt, several recent systems use multiple agents with distinct roles or perspectives. Some frameworks decompose a claim into smaller units and verify each one independently. Others emulate debate, critique, or consensus-building among specialized agents. LoCal-style decomposition methods explicitly represent logical and causal structure, while Delphi-inspired systems combine the outputs of multiple reasoning personas into a final collective judgment. These ideas are particularly appealing for misinformation detection because many claims are multi-part, causally loaded, or rhetorically framed in ways that benefit from diversified reasoning.

Our work draws on all of these threads, but differs in how it integrates them. TRUST Agents is not only a multi-agent verifier. It is a structured end-to-end framework that includes claim extraction, hybrid retrieval, evidence-grounded verification, explanation generation, claim decomposition, verifier diversity, and logic-aware aggregation. Rather than using multiple agents only for debate, we organize them into a pipeline where each agent has a defined responsibility, allowing the overall system to function as an interpretable fact-checking architecture.

\section{Method}
\subsection{System Overview}
TRUST Agents is designed as a collaborative multi-agent framework for end-to-end fact verification. The architecture contains two execution modes. The first is a baseline pipeline designed for direct claim verification. The second is a research-enhanced pipeline intended for compositional claims and more structured reasoning.

The baseline pipeline consists of four core modules. A Claim Extractor Agent identifies factual claims from input text. An Evidence Retrieval Agent gathers passages from a trusted corpus using hybrid search. A Verifier Agent compares claims against retrieved evidence and produces verdicts with confidence scores. An Explainer Agent converts those outputs into transparent reports with evidence citations.

The research-enhanced pipeline adds three reasoning-oriented modules. A Decomposer Agent breaks complex claims into atomic sub-claims and returns logical structure. A Delphi Multi-Agent Jury verifies each atomic claim using multiple specialized verifier personas. A Logic Aggregator then combines the atomic outputs according to the formula returned by the decomposer. The result is a system that can reason over compound claims more explicitly than the baseline pipeline.

\begin{figure}[H]
    \centering
    \includegraphics[width=0.88\textwidth]{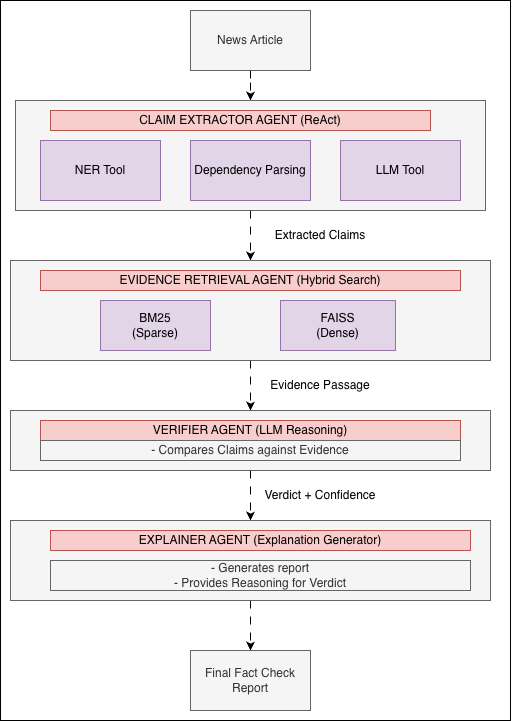}
    \caption{Baseline TRUST Agents architecture. A news article is processed through claim extraction, hybrid evidence retrieval, verification, and explanation generation to produce a final fact-check report.}
    \label{fig:baseline_architecture}
\end{figure}

\begin{figure}[H]
    \centering
    \includegraphics[width=0.78\textwidth]{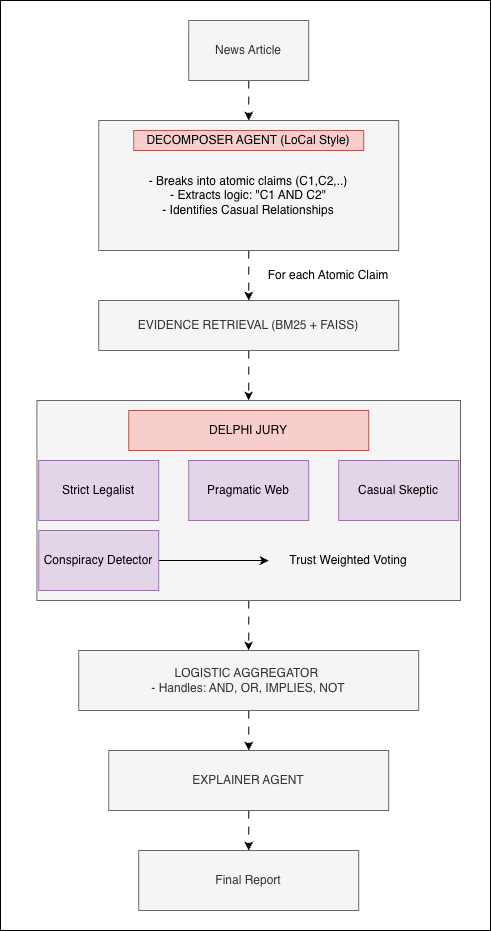}
    \caption{Research extension of TRUST Agents. The system first decomposes the input into atomic claims, verifies each atomic claim using a multi-persona jury, aggregates the results with a logic-aware module, and produces a final report.}
    \label{fig:research_architecture}
\end{figure}

Figure~\ref{fig:baseline_architecture} illustrates the baseline architecture. Figure~\ref{fig:research_architecture} illustrates the research extension.

\subsection{Baseline Pipeline}
\subsubsection{Claim Extractor Agent}
The claim extraction stage is responsible for identifying verifiable assertions from raw input text. This stage is necessary because a news article or user-provided paragraph often contains a mix of background context, interpretation, rhetorical commentary, and factual claims. Treating the entire text as one claim would reduce precision and make downstream retrieval noisier.

The Claim Extractor Agent combines three complementary extraction tools. The first tool uses named entity recognition through spaCy's \texttt{en\_core\_web\_sm} model. The intuition is that many factual claims mention concrete entities such as people, organizations, locations, dates, events, or monetary quantities. Sentences lacking such content are less likely to be directly verifiable and can often be filtered out. The second tool uses dependency parsing to identify subject--verb--object patterns and to detect whether a sentence expresses a factual relation. We use a curated set of claim-bearing verbs such as \texttt{say}, \texttt{claim}, \texttt{report}, \texttt{show}, \texttt{estimate}, and \texttt{predict}. A sentence containing a meaningful subject, a claim-related verb, and a plausible object is treated as a strong candidate factual statement. The third tool uses an LLM prompt to extract claims that are implicit, semantically complex, or difficult to capture through rule-based parsing alone. The model is instructed to return a clean JSON list of distinct factual claims while excluding opinions and speculative language.

These tools are coordinated using a ReAct-style control loop. The agent keeps track of which tools have already been used, merges overlapping outputs, removes duplicates, and returns a final claim list. This hybrid setup allows the extractor to capture both explicit claims and more nuanced propositions without relying exclusively on either shallow parsing or free-form generation.

\subsubsection{Evidence Retrieval Agent}
Once claims are extracted, the next step is to retrieve relevant passages from a trusted evidence corpus. Retrieval quality is critical because all downstream reasoning depends on the evidence available to the verifier. A strong verifier cannot compensate for missing or irrelevant evidence.

The Evidence Retrieval Agent uses a hybrid sparse-plus-dense retrieval strategy. Source documents are first converted to plain text and split into overlapping passages. In our implementation, chunks are typically around 160 words with an overlap of approximately 20 words. This chunking policy helps preserve context while allowing fine-grained retrieval. Each passage is then inserted into two indices. The first is a BM25 sparse index, implemented through Pyserini, which captures lexical overlap and exact phrase matches. The second is a FAISS dense index built over sentence-transformer embeddings, which supports semantic retrieval even when the claim and the evidence use different wording.

Given a claim, the system computes both BM25 scores and dense similarity scores for candidate passages. These are combined into a weighted hybrid score. A representative formulation is
\[
\mathrm{score}_{\text{hybrid}} = 0.6 \cdot \mathrm{score}_{\text{BM25}} + 0.4 \cdot \mathrm{score}_{\text{dense}}.
\]
After the top candidates are produced, Maximal Marginal Relevance is applied to encourage diversity in the selected evidence set. This is important because the highest-scoring passages are often redundant paraphrases of one another, and a more diverse evidence bundle is usually more informative for verification.

The retrieval agent is exposed through a set of tools for index building, passage search, specific passage lookup, and corpus inspection. This tool interface makes it easier to debug retrieval failures and to replace components in future work.

\subsubsection{Verifier Agent}
The Verifier Agent receives a claim and a set of retrieved evidence passages, and is responsible for determining whether the evidence supports, contradicts, or is insufficient to judge the claim. Instead of producing a single opaque prediction, the verifier operates in stages so that its intermediate reasoning can be inspected.

For each evidence passage, the agent prompts an LLM to compare the passage with the claim. The output is a structured JSON object containing a label indicating support, contradiction, or insufficiency; a confidence score; and a short list of key reasoning points. This transforms claim verification into a sequence of explicit claim--evidence comparisons rather than a single hidden inference.

The passage-level outputs are then aggregated. If multiple high-quality passages agree that the claim is supported, confidence rises. If multiple passages contradict the claim, the final score shifts in the opposite direction. If the evidence is mixed, sparse, or weak, the system lowers confidence and may abstain. This aggregation step also takes evidence quality into account, so that a low-quality or weakly matching passage does not influence the final verdict as much as a highly relevant one.

The final verdict is mapped into one of three labels: \texttt{true}, \texttt{false}, or \texttt{uncertain}. Confidence is calibrated into the interval $[0,1]$. This design intentionally allows abstention, since forcing a binary decision under poor evidence often produces misleading outputs.

\subsubsection{Explainer Agent}
A central design goal of TRUST Agents is interpretability. The Explainer Agent takes the intermediate outputs from the verifier and converts them into a structured, human-readable report.

The explainer first generates a concise summary of the verification process. It then formats the evidence passages into numbered citations, including source references where available. Finally, it produces a more detailed narrative explaining how the evidence supports or contradicts the claim and why the final verdict was chosen. The explanation is grounded in the retrieved passages rather than generated as an unsupported free-form rationale. The final report contains the claim, verdict, confidence score, summary, explanation, and citations.

\subsection{Research-Enhanced Pipeline}
The baseline pipeline provides modular evidence-grounded fact verification, but it still has two limitations. First, it treats each extracted claim as a flat statement even when the claim contains multiple propositions connected by logic or causality. Second, it relies on a single verifier, which means the result may reflect the biases or blind spots of one particular reasoning style. To address these limitations, we introduce a research-enhanced pipeline.

\subsubsection{Decomposer Agent}
The Decomposer Agent is inspired by LoCal-style logical claim decomposition. Its purpose is to transform a complex claim into a set of atomic claims and an explicit logical structure. This is important because many real-world claims are not simple propositions. They may contain conjunctions such as ``X happened and Y happened,'' disjunctions such as ``either A or B caused C,'' implications such as ``if policy P was implemented, outcome Q followed,'' or hidden causal assumptions.

The decomposer prompts an LLM to return a JSON object containing the list of atomic claims, a logical formula connecting them, a list of causal edges if applicable, and a complexity score in the range $[0,1]$. For example, a claim such as ``The policy reduced unemployment and increased wages in 2024'' may be decomposed into one atomic claim about unemployment, one about wages, and one about the relevant time period. The formula would then specify that the full claim is true only if all of the required parts hold.

If the model fails to produce a valid decomposition, the system falls back to treating the entire claim as one atomic unit. This keeps the pipeline robust during large-scale experiments.

\subsubsection{Delphi Multi-Agent Jury}
After decomposition, each atomic claim is passed to a Delphi-inspired verifier jury. The goal is to reduce reliance on one reasoning pattern by allowing multiple specialized verifier personas to assess the same evidence from different perspectives.

The \textit{Strict Legalist} persona is conservative and prefers highly credible sources such as government institutions, major news outlets, and academic organizations. It is reluctant to commit when evidence is weak or when there are not enough independent confirmations. The \textit{Open Web Pragmatist} persona is more flexible and is willing to consider a wider range of sources, provided that the content is coherent and relevant. The \textit{Causal Skeptic} persona focuses on temporal order, numeric consistency, and causal validity, making it especially useful for claims that infer causation from correlation. The \textit{Conspiracy Detector} persona focuses on claims that rely on cherry-picked evidence, sweeping hidden-agent assumptions, or extraordinary unsupported conclusions.

Each persona produces a verdict, a confidence score, and a brief explanation. These persona outputs are not simply averaged. Instead, we compute a trust score for each persona based on the quality of the retrieved evidence, the persona's internal confidence, and whether the reasoning completed cleanly without failure. A typical trust score is computed as
\[
\mathrm{trust}_p = 0.4 \cdot \overline{s}_{\text{hybrid}} + 0.4 \cdot c_p + 0.2 \cdot \mathbb{1}_{\text{no error}},
\]
where $\overline{s}_{\text{hybrid}}$ is the average hybrid retrieval score, $c_p$ is the persona confidence, and $\mathbb{1}_{\text{no error}}$ indicates successful completion. The final label for the atomic claim is then determined through trust-weighted voting:
\[
\mathrm{Vote}(v)=\sum_{p \in \mathcal{P}} \mathrm{trust}_p \cdot c_p \cdot \mathbb{1}[\hat{y}_p=v].
\]
This design allows more reliable or better-grounded personas to contribute more strongly to the final decision.

\subsubsection{Logic Aggregator}
Once the atomic claims have been verified, the Logic Aggregator combines them according to the formula returned by the decomposer. The aggregator maps verdicts such as \texttt{true} and \texttt{supported} into Boolean-like positive states, and verdicts such as \texttt{false} and \texttt{contradicted} into Boolean-like negative states. \texttt{Uncertain} is represented as an unknown state.

The logical formula is then evaluated. Conjunctions require all relevant atomic claims to hold. Disjunctions require at least one supporting atomic claim. Negation flips the polarity of a claim. Implication is converted into an equivalent Boolean form such as $(\neg A)\vee B$. If one or more atomic claims remain uncertain, the system may propagate that uncertainty to the overall result, depending on the formula. When logic parsing fails due to malformed output or unsupported syntax, the system falls back to a simpler majority-vote strategy over the atomic claims.

This aggregation step is one of the main differences between the research-enhanced pipeline and standard multi-agent verification systems. Instead of just collecting multiple agent opinions, it forces the final prediction to respect the internal compositional structure of the claim.

\subsection{System Orchestration}
The full framework supports two orchestration modes. In baseline mode, the system runs claim extraction, retrieval, verification, and explanation sequentially. In research mode, it runs decomposition first, then retrieves evidence for each atomic claim, invokes the verifier jury, aggregates the outputs through the logic module, and finally generates a structured explanation. Both orchestrators log intermediate states, making it possible to inspect errors in retrieval, decomposition, voting, or aggregation during analysis.

\section{Experimental Setup}
\subsection{Dataset}
We evaluate the system on the LIAR benchmark, which contains 12{,}836 short political statements annotated with six truthfulness labels. The dataset includes additional metadata such as the speaker, political party, subject, and context in which the statement was made. The benchmark is useful because it contains concise, real-world claims and has become a common reference point for truthfulness prediction models.

For evaluation, we adopt a binary mapping of the six labels. The labels \texttt{true}, \texttt{mostly true}, and \texttt{half true} are mapped to the positive class, while \texttt{false}, \texttt{pants on fire}, and \texttt{barely true} are mapped to the negative class. This simplified setup enables comparison with supervised baselines. However, because TRUST Agents can also output \texttt{uncertain}, the evaluation is necessarily approximate and requires extra discussion.

\subsection{Baselines}
We compare TRUST Agents against three strong baselines. The first is fine-tuned BERT. The second is fine-tuned RoBERTa. These two models are standard encoder-only baselines for LIAR-style classification tasks. The third is a zero-shot LLM baseline that judges claims directly without the full multi-agent retrieval-and-reasoning pipeline.

BERT and RoBERTa are fine-tuned on the LIAR training split using a learning rate of $2\times10^{-5}$, batch size 16, and two epochs. The zero-shot LLM baseline is evaluated directly over the benchmark claims.

\subsection{Implementation Details}
The system is implemented using spaCy for named entity recognition and dependency parsing, Pyserini for BM25 retrieval, FAISS for dense retrieval, and sentence-transformer embeddings for semantic passage indexing. LangChain and LangGraph are used to implement the multi-agent orchestration and ReAct-style tool use. HuggingFace Transformers is used to fine-tune the BERT and RoBERTa baselines.

The evidence retrieval system indexes a subset of trusted sources, including Wikipedia pages and selected news or government documents. LLM calls are run with a low temperature to reduce output variance. Experiments were conducted on a machine with an 8-core CPU and 32 GB RAM.

\subsection{Evaluation Protocol}
We report accuracy and macro-F1 as the main evaluation metrics. Accuracy gives a general measure of correctness, while macro-F1 is more informative under class imbalance and is less dominated by the majority class.

Because TRUST Agents can output \texttt{uncertain}, we evaluate it under two mappings. In the pessimistic setting, \texttt{uncertain} is mapped to \texttt{false}. In the optimistic setting, \texttt{uncertain} is mapped to \texttt{true}. Neither mapping is ideal, but they help bracket the system's behavior under binary evaluation. We also report the raw proportion of abstentions during analysis.

\section{Results}
\subsection{Supervised Baselines}
Table~\ref{tab:supervised-results} shows the performance of fine-tuned BERT, fine-tuned RoBERTa, and the zero-shot LLM baseline.

\begin{table}[H]
    \centering
    \begin{tabular}{lcc}
        \toprule
        \textbf{Model} & \textbf{Accuracy} & \textbf{Macro-F1} \\
        \midrule
        BERT (fine-tuned) & 0.652 & 0.726 \\
        RoBERTa (fine-tuned) & 0.641 & 0.726 \\
        GPT-4.1-nano (zero-shot) & 0.580 & 0.528 \\
        \bottomrule
    \end{tabular}
    \caption{Performance of supervised and zero-shot baselines on the LIAR test set.}
    \label{tab:supervised-results}
\end{table}

The supervised encoder baselines remain substantially stronger than the multi-agent system in raw benchmark accuracy and macro-F1. This is not surprising because they are trained directly to predict the benchmark labels. By contrast, TRUST Agents is designed as an evidence-grounded verifier that can abstain when evidence is weak or incomplete.

\subsection{Baseline and Research Pipelines}
Table~\ref{tab:pipeline-results} summarizes the performance of the original baseline pipeline and the research-enhanced pipeline.

\begin{table}[H]
    \centering
    \begin{tabular}{llcc}
        \toprule
        \textbf{Pipeline} & \textbf{Uncertainty Mapping} & \textbf{Accuracy} & \textbf{Macro-F1} \\
        \midrule
        Original & none & $\approx 0.19$ & $\approx 0.19$ \\
        Original & uncertain $\rightarrow$ false & 0.485 & 0.428 \\
        Original & uncertain $\rightarrow$ true & 0.520 & 0.434 \\
        Research & uncertain $\rightarrow$ false & 0.495 & 0.363 \\
        Research & uncertain $\rightarrow$ true & 0.520 & 0.444 \\
        \bottomrule
    \end{tabular}
    \caption{Performance of the original and research TRUST pipelines under different uncertainty mappings.}
    \label{tab:pipeline-results}
\end{table}

The most important pattern is the high rate of abstention. The original pipeline returns \texttt{uncertain} for roughly 70\% of cases, while the research pipeline increases this to approximately 82\%. Under standard benchmark metrics, this behavior strongly limits measured performance. However, this conservatism also reflects the system's reluctance to overstate confidence when the evidence is weak or contradictory.

The research pipeline yields a modest improvement in macro-F1 under the optimistic uncertainty mapping, increasing from 0.434 to 0.444. More importantly, it provides significantly richer internal structure. The system exposes atomic claims, logical formulas, persona-level judgments, trust-weighted voting behavior, and logic-aware aggregation. These signals are useful for understanding why the system reached a decision and where it became uncertain.

\section{Discussion}
The experimental results highlight a central tension in evidence-grounded fact-checking. A system that is cautious and willing to abstain may appear weak under binary benchmark metrics, even though abstention is often the correct behavior in realistic settings. This is particularly relevant for misinformation detection, where overconfident but unsupported judgments can cause harm.

The strongest qualitative benefit of TRUST Agents lies in its transparency. Unlike a direct classifier, it reveals the steps of the verification process. The baseline pipeline makes it possible to inspect extracted claims, retrieved evidence passages, and passage-level support judgments. The research-enhanced pipeline goes further by making claim composition explicit. It decomposes complex inputs into smaller factual units, shows how different verifier personas interpret the evidence, and combines those interpretations according to formal logical structure. This makes the system much more useful for debugging, analysis, and human oversight.

At the same time, the results make it clear that reasoning improvements alone are not enough. Retrieval quality remains the main bottleneck. If the retrieval agent fails to find relevant evidence, even the best decomposition and voting mechanism will often produce \texttt{uncertain}. Many LIAR claims refer to political context, speaker history, or time-sensitive events that are not well covered by a limited evidence corpus. As a result, abstention becomes the default safe behavior.

The research-enhanced pipeline also introduces a cost-benefit tradeoff. Decomposition, multi-persona verification, and logic-aware aggregation improve interpretability and compositional reasoning, but they also increase latency and system complexity. This suggests that future versions of the framework may benefit from adaptive orchestration. A simple claim with strong evidence may only need the baseline pipeline, while a complex or ambiguous claim could trigger the full research pipeline.

\section{Limitations}
The current system has several important limitations. The first is the high abstention rate. Although abstention is often safer than overconfidence, it reduces benchmark performance and reveals that the system still lacks sufficient evidence coverage and calibration. The second limitation is the restricted evidence corpus. The current retrieval setup relies primarily on Wikipedia and a selected set of trusted documents, which is not enough to cover the diversity of real-world misinformation claims. Claims about very recent events, niche domains, or international contexts are especially difficult to verify.

A third limitation is model capacity. To manage computational cost and latency, the pipeline uses relatively small LLM variants for extraction, decomposition, verification, and explanation. These models are more affordable but also more likely to default to uncertainty or to produce brittle outputs when evidence is incomplete. A fourth limitation lies in the evaluation protocol. TRUST Agents naturally operates as a three-way system with \texttt{true}, \texttt{false}, and \texttt{uncertain} outputs, but LIAR is essentially evaluated in a binary setting. Mapping \texttt{uncertain} to one of the binary classes is only an approximation and does not fully capture the value of calibrated abstention. Finally, LIAR itself is limited to short U.S. political statements and therefore does not test longer-form misinformation, domain diversity, or multilingual robustness.

\section{Broader Impact and Ethics}
Automated fact-checking systems can provide substantial public benefit when they are used as assistive tools for journalists, moderators, researchers, or end users. A system that can retrieve evidence quickly, summarize relevant passages, and expose reasoning structure can reduce the burden on human fact-checkers and help prioritize claims for deeper investigation. However, these systems also introduce risks. If an automated verifier appears more reliable than it actually is, users may over-trust its conclusions. This risk becomes more serious when the system operates in politically sensitive or high-stakes domains.

For this reason, we emphasize evidence grounding, explanation transparency, and uncertainty awareness throughout the system design. TRUST Agents is not intended to replace human judgment. Instead, it is designed to support human-in-the-loop verification by surfacing claims, evidence, disagreements, and reasoning traces. Any practical deployment of such a system should preserve human oversight, expose uncertainty clearly, and allow users to inspect the evidence behind each output.

\section{Conclusion}
This paper presented TRUST Agents, a collaborative multi-agent framework for fake news detection, evidence-grounded claim verification, and logic-aware reasoning. The baseline system integrates claim extraction, hybrid retrieval, verification, and explanation generation into a modular end-to-end pipeline. The research-enhanced variant extends this architecture with claim decomposition, a Delphi-inspired verifier jury, and logic-aware aggregation for compound claims.

Although the system does not yet surpass supervised encoder baselines on raw benchmark metrics, it offers substantial advantages in interpretability, modularity, and structured reasoning. The research-enhanced pipeline is especially valuable for handling multi-part claims and for exposing the internal disagreement structure of the verification process. The experiments also show that the main obstacles are not only reasoning quality, but also retrieval coverage and uncertainty calibration. These findings suggest that future progress in trustworthy fact-checking will depend on better evidence retrieval, stronger uncertainty-aware evaluation, and adaptive reasoning pipelines that invoke more complex modules only when they are truly needed.

\section*{Code Availability}
The implementation of TRUST Agents, including the baseline pipeline, research pipeline, retrieval code, and evaluation scripts, is available at:
\begin{center}
\url{https://github.com/GautamaShastry/news_agent}
\end{center}

\end{document}